\documentclass[runningheads]{llncs}

\usepackage{graphicx}
\usepackage{booktabs}
\usepackage{amsmath}
\usepackage{hyperref}

\begin{document}

\title{Do Diabetic Foot Ulcer Segmentation Models Generalize?\\
A Cross-Dataset Benchmark of CNN and Transformer Architectures}
\titlerunning{Cross-Dataset DFU Segmentation Benchmark}

\author{Abderrahmane Benfatah}
\authorrunning{A. Benfatah}
\institute{King Saud University, Riyadh, Saudi Arabia \\
\email{444106928@student.ksu.edu.sa}}

\maketitle

% =============================================================
\begin{abstract}
Deep learning models for diabetic foot ulcer (DFU) segmentation routinely
report high accuracy, but they are almost always trained and tested on the
same dataset, leaving their behaviour on data from a different clinical
source largely unmeasured. We benchmark three representative segmentation
architectures---U-Net and DeepLabV3+ (convolutional) and SegFormer-B2
(Transformer)---under an identical, leakage-screened protocol: training on
the combined FUSeg/AZH wound data and evaluating, without fine-tuning, on
two independent external datasets (DFUC2022 and Medetec). All models achieve
strong in-domain performance (Dice $0.80$--$0.83$) but degrade substantially
across datasets. The degradation is, however, architecture-dependent:
SegFormer-B2 generalizes best on both external sets
(DFUC2022 Dice $0.557$, Medetec Dice $0.786$), outperforming both
convolutional models, while the more complex DeepLabV3+ generalizes worse
than the simpler U-Net. Per-image failure analysis on $2{,}160$ images
across both external test sets confirms that SegFormer-B2 produces the
fewest catastrophic failures on DFUC2022 (31.1\%), compared with U-Net
(38.5\%) and DeepLabV3+ (43.0\%). The consistent ranking across two
independent external sources, confirmed by Wilcoxon signed-rank tests
($p < 0.001$ on both datasets), indicates that architecture family, not
model complexity, drives cross-hospital generalization.

\end{abstract}

% =============================================================
\section{Introduction}

Diabetic foot ulcers (DFU) are a severe complication of diabetes mellitus and
a leading cause of non-traumatic lower-limb amputation. The lifetime risk of
developing a foot ulcer among people with diabetes is estimated at
19--34\%, and roughly one in five patients who develop a DFU ultimately
requires a lower-extremity amputation, with reported five-year mortality of
50--70\%~\cite{armstrong2023}. Early detection and continuous monitoring of
wound size are therefore clinically important, and automatic segmentation of
the wound region from photographs is a key enabling step for objective area
measurement and longitudinal tracking.

Convolutional neural networks (CNNs), and in particular the U-Net
architecture~\cite{unet}, have become the standard approach for wound
segmentation and achieve high accuracy when trained and evaluated on a single
dataset~\cite{fusegnet}. These scores, however, are almost always reported
\emph{in-domain}, on test images drawn from the same source as the training
data.

Clinical deployment, by contrast, requires models to perform on images from
hospitals, cameras, and patient populations not seen during training. Prior
work has shown that DFU segmentation models lose substantial accuracy when
evaluated across datasets~\cite{lucho2024}, but this analysis was limited to
convolutional and prompt-based (SAM) models. While recent work has applied
Transformer architectures including SegFormer to DFU segmentation, these
studies focused on in-domain performance on a single
dataset~\cite{segformer_ensemble,ulcermtl}; to the best of our knowledge, no
prior study has conducted a \emph{leakage-screened} cross-dataset evaluation
of a modern Transformer against CNN baselines for DFU segmentation, nor
provided a per-image failure analysis confirming the robustness of the ranking
across independent test sets.

In this work we benchmark three architectures spanning two families---U-Net
and DeepLabV3+ (CNN) and SegFormer-B2 (Transformer)---under one identical,
leakage-screened protocol, training on the combined FUSeg/AZH data and testing
without fine-tuning on two independent external datasets. Our contributions
are:
\begin{itemize}
\item A controlled cross-dataset comparison of CNN \emph{and} Transformer
segmentation models for DFU, which prior cross-dataset studies did not
include.
\item A leakage-screened evaluation protocol, after auditing seven public
wound datasets for image-level overlap.
\item Validation on two independent external test sets (DFUC2022 and
Medetec), confirming that architecture family---not complexity---drives
cross-hospital generalization.
\item A per-image failure analysis across $2{,}160$ images, with Wilcoxon
signed-rank tests confirming that SegFormer-B2 significantly outperforms
both CNN baselines ($p < 0.001$) on both external datasets.
\end{itemize}

% =============================================================
\section{Related Work}

\textbf{DFU segmentation.}
Wound and DFU segmentation has progressed from early encoder--decoder
networks to U-Net variants enhanced with transfer learning and attention
mechanisms. The FUSeg challenge~\cite{fuseg} standardized a benchmark of
$1{,}210$ annotated foot-ulcer images, and challenge-winning approaches such
as FUSegNet~\cite{fusegnet}, which augments a U-Net-style decoder with
parallel squeeze-and-excitation modules, report Dice scores near $0.89$.
More recently, Transformer architectures have been applied to DFU
segmentation: SegFormer-based ensembles~\cite{segformer_ensemble} and
multi-task learning frameworks using SegFormer-B2 as a
backbone~\cite{ulcermtl} have shown promising in-domain results. These
methods are, however, optimized and evaluated within individual datasets;
their cross-hospital generalization remains unmeasured.

\textbf{Transformers and foundation models.}
Vision Transformers and foundation models such as MedSAM~\cite{medsam} have
recently been applied to medical and wound segmentation, motivated by their
ability to capture global context. Their evaluation in the wound domain has
nonetheless remained largely in-domain or prompt-based rather than focused on
cross-dataset robustness.

\textbf{Cross-dataset generalization.}
The closest work to ours, Lucho et al.~\cite{lucho2024}, benchmarked
convolutional models and SAM across the FUSeg and DFUC datasets and reported
large performance drops when training and testing on different datasets. That
study did not evaluate a modern Transformer segmentation model, nor did it
screen the datasets for cross-source image overlap. While SegFormer has since
been applied to DFU segmentation for in-domain tasks~\cite{segformer_ensemble,ulcermtl},
its cross-dataset generalization has not been measured, and no prior study has
applied a leakage-screening protocol before evaluation. To the best of our
knowledge, this is the \emph{first leakage-screened, cross-dataset comparison}
of CNN and Transformer architectures for DFU segmentation. We additionally
validate on a second independent test set (Medetec) and confirm all rankings
with per-image statistical tests.

% =============================================================
\section{Methods}

\subsection{Datasets}
We use four public DFU/wound segmentation datasets. FUSeg (the Foot Ulcer
Segmentation Challenge) contains $1{,}210$ foot-ulcer images from 889
patients with expert annotations~\cite{fuseg}. The AZH Wound Care Center
dataset provides 1{,}010 additional chronic-wound images with segmentation
masks (overlapping with the FUSeg training split; treated as a single combined
source after leakage screening). DFUC2022 is an independent DFU dataset
collected at a different clinical site, used as external test set~\#1
($2{,}000$ images). Medetec (160 images, $224\times224$~px) is a second
independent wound dataset with zero image overlap with any training source,
used as external test set~\#2.

\subsection{Dataset Integrity and Leakage Screening}
Because several public wound datasets are repackaged or aggregated versions of
one another, we first audited all available sources for image-level overlap.
Every image was fingerprinted using both an exact decoded-pixel hash (MD5) and
a perceptual hash (dHash), and all dataset pairs were compared; flagged pairs
were additionally verified by visual inspection to reject false positives.
The audit revealed that one aggregated folder was almost entirely composed of
DFUC2022 images, and that FUSeg and AZH share a large fraction of identical
images. We therefore treat FUSeg and AZH as a single combined training source
and confirmed that the training set (FUSeg/AZH) and both external test sets
contain no meaningful overlap. Note: Medetec segmentation masks encode wound
labels as $\{0,\,1\}$ pixel values rather than the conventional $\{0,\,255\}$;
masks were thresholded at $>0$ accordingly.

\subsection{Models}
We evaluate three architectures spanning two families. As convolutional
baselines we use \textbf{U-Net}~\cite{unet} and
\textbf{DeepLabV3+}~\cite{deeplab}, both with a ResNet-34 encoder.
As a Transformer model we use \textbf{SegFormer-B2}~\cite{segformer}
(MiT-B2 encoder). All encoders are initialized with ImageNet-pretrained
weights and implemented in
\texttt{segmentation\_models\_pytorch}~\cite{deeplabv3plus_repo}.

\subsection{Training and Evaluation Protocol}
All models share an identical protocol. Inputs are resized to
$256\times256$. We train with a combined Dice and binary cross-entropy loss
for 30 epochs. Convolutional models use the Adam optimizer with a learning
rate of $1\times10^{-3}$; SegFormer uses AdamW with a learning rate of
$6\times10^{-5}$, following common practice for Transformers. Standard data
augmentation (horizontal flip, brightness/contrast jitter, and affine
transforms) is applied during training only. The checkpoint with the best
validation Dice is retained. We report the Dice coefficient (F1) and
Intersection-over-Union (IoU). DFUC2022 results are averaged over three
random seeds (42, 123, 2024) and reported as mean~$\pm$~std; Medetec results
are from seed 42 (single run, 160 images). The \emph{generalization gap} is
defined as in-domain Dice minus cross-dataset Dice. Statistical significance
of pairwise per-image Dice differences is assessed using the Wilcoxon
signed-rank test (one-sided, $\alpha = 0.05$).

\subsection{Reproducibility}
All experiments were conducted using PyTorch~2.1 and
\texttt{segmentation\_models\_pytorch}~0.3.3 on an NVIDIA RTX~4050 Laptop
GPU (6\,GB VRAM, CUDA~12.1). Each model was trained for 30 epochs
(batch size~8, input $256\times256$); wall-clock training time was
approximately 20 minutes per run. Training scripts, leakage-audit scripts,
and evaluation scripts are publicly available at:
\url{https://github.com/ben-fatah/ILTIAM-DFU-Benchmark}

\subsection{Threats to Validity}
Several limitations should be noted. First, only one Transformer architecture
(SegFormer-B2) was evaluated; other variants such as Swin-UNet or MedSAM may
behave differently. Second, both external test sets are limited in size
(DFUC2022: $2{,}000$ images; Medetec: 160 images), and conclusions may not
extend to larger or more diverse clinical populations. Third, all datasets
contain diabetic foot ulcers specifically; results may not generalise to other
wound categories such as pressure ulcers or venous leg ulcers. Fourth, all
models were trained at $256\times256$ due to GPU memory constraints;
SegFormer-B2 was originally designed for $512\times512$ inputs, so our
reported cross-dataset Dice for SegFormer may be \emph{conservative}---its
advantage over CNNs could be larger at higher resolution. Fifth, no learning
rate scheduler was used, ensuring a fair apples-to-apples comparison across
architectures but potentially leaving performance gains on the table.

% =============================================================
\section{Results and Discussion}

\subsection{In-domain Performance}
Table~\ref{tab:results} reports performance for all three architectures.
On the held-out FUSeg validation set---data from the same distribution as
training---all models perform comparably and strongly. SegFormer-B2 attains
the highest in-domain Dice ($0.834\pm0.003$), followed by U-Net
($0.817\pm0.007$) and DeepLabV3+ ($0.802\pm0.012$). The narrow spread
($\approx0.03$ Dice) indicates that architecture choice has only a marginal
effect within a single source.

\subsection{Cross-dataset Generalization}

\noindent\textbf{DFUC2022 (external \#1).}
SegFormer-B2 generalizes best (Dice $0.557\pm0.003$, gap 0.277), whereas
U-Net falls to $0.501\pm0.029$ (gap 0.316) and DeepLabV3+ to
$0.489\pm0.019$ (gap 0.313). Beyond mean accuracy, SegFormer is also
markedly more \emph{stable} across seeds: its cross-dataset standard
deviation ($\pm0.003$) is an order of magnitude smaller than U-Net's
($\pm0.029$), indicating that the convolutional models' cross-hospital
performance is partly dependent on random initialization, whereas the
Transformer's is not.

\noindent\textbf{Medetec (external \#2).}
The same ranking holds: SegFormer-B2 leads (Dice 0.786, IoU 0.682),
followed by U-Net (Dice 0.737, IoU 0.613) and DeepLabV3+
(Dice 0.730, IoU 0.607). Absolute scores are higher on Medetec because
the domain shift is imaging-protocol only (same DFU wound type), whereas
DFUC2022 combines imaging and clinical-site shifts.

\begin{table}[t]
\centering
\caption{In-domain versus two independent cross-dataset test sets.
DFUC2022 results are mean~$\pm$~std over three seeds (42, 123, 2024).
Medetec results are single-seed (seed 42). Bold = best per column.
Gap = in-domain Dice $-$ DFUC2022 Dice.}
\label{tab:results}
\setlength{\tabcolsep}{3.5pt}
\begin{tabular}{llcccccc}
\toprule
Model & Family & FUSeg & DFUC2022 & DFUC2022 & Gap & Medetec & Medetec \\
      &        & Dice  & Dice     & IoU      &     & Dice    & IoU \\
\midrule
U-Net (ResNet34)      & CNN         & $0.817\pm0.007$ & $0.501\pm0.029$ & 0.414 & 0.316 & 0.737 & 0.613 \\
DeepLabV3+ (ResNet34) & CNN         & $0.802\pm0.012$ & $0.489\pm0.019$ & 0.398 & 0.313 & 0.730 & 0.607 \\
SegFormer-B2          & Transformer & $\mathbf{0.834\pm0.003}$ & $\mathbf{0.557\pm0.003}$ & \textbf{0.458} & \textbf{0.277} & \textbf{0.786} & \textbf{0.682} \\
\bottomrule
\end{tabular}
\end{table}

\subsection{Failure Analysis}
To move beyond aggregate metrics, we compute per-image Dice scores for all
three models on both external test sets and identify failure cases, defined
as images with Dice $< 0.4$ (Table~\ref{tab:failure}).

\begin{figure}[t]
\centering
\includegraphics[width=\textwidth]{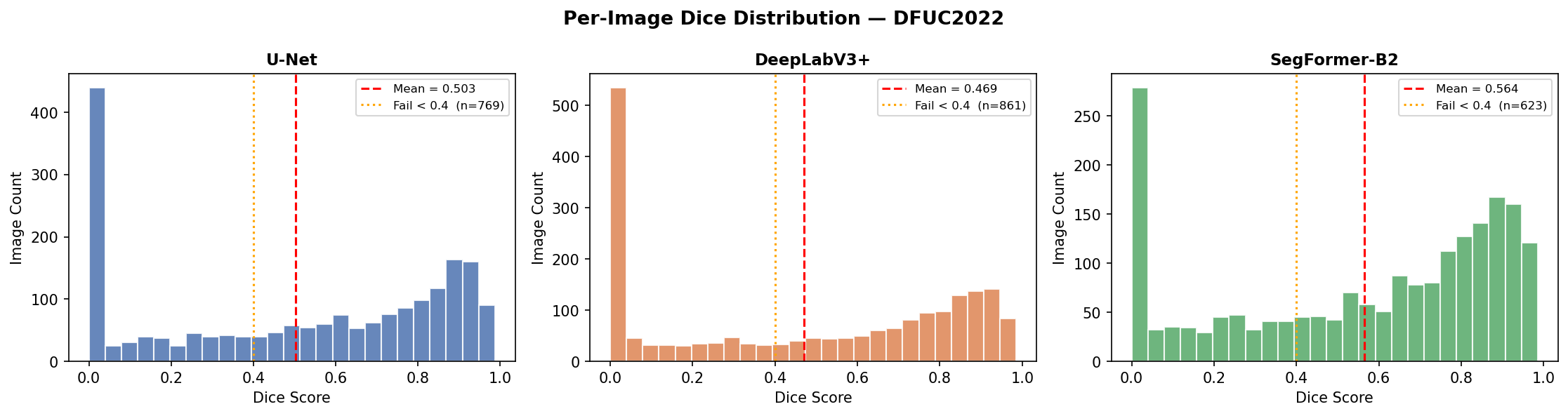}
\caption{Per-image Dice distributions on DFUC2022. SegFormer-B2 shows a
rightward shift and reduced left tail relative to both CNN models,
corresponding to fewer images below the failure threshold of 0.4
(orange dashed line).}
\label{fig:hist}
\end{figure}

\noindent\textbf{DFUC2022 failure rates.}
SegFormer-B2 produces the fewest failures on DFUC2022 (623 images, 31.1\%),
followed by U-Net (769, 38.5\%) and DeepLabV3+ (861, 43.0\%). The gap
between SegFormer-B2 and DeepLabV3+ corresponds to 238 additional images
segmented above the failure threshold---confirming that the Transformer
advantage observed in mean Dice reflects a consistent per-image improvement
rather than a small number of high-scoring outliers (Fig.~\ref{fig:hist}).

\noindent\textbf{Medetec failure rates and a nuance.}
On Medetec, absolute failure rates are low for all models. The count-based
ranking reverses: SegFormer-B2 registers the most failures by image count
(14, 8.8\%) compared with U-Net (11, 6.9\%) and DeepLabV3+ (13, 8.1\%),
yet retains the highest mean Dice (0.786). This apparent contradiction is
resolved by examining the score distributions: when SegFormer-B2 fails, it
fails \emph{gracefully}---its failures cluster just below the 0.4 threshold---whereas
CNN failures include a heavier catastrophic tail (Dice $\approx 0$), as
visible in the Dice histograms (Fig.~\ref{fig:hist}). SegFormer-B2 therefore
fails less severely even when it does fail, and mean Dice remains the more
informative summary statistic on Medetec.

\begin{table}[t]
\centering
\caption{Per-image failure rates (Dice $< 0.4$) on both external test sets.
DFUC2022: $n=2{,}000$; Medetec: $n=160$. Bold = fewest failures.}
\label{tab:failure}
\begin{tabular}{llcccc}
\toprule
Model & Family & \multicolumn{2}{c}{DFUC2022} & \multicolumn{2}{c}{Medetec} \\
      &        & Failed & Fail\% & Failed & Fail\% \\
\midrule
U-Net (ResNet34)      & CNN         & 769 & 38.5 & \textbf{11} & \textbf{6.9} \\
DeepLabV3+ (ResNet34) & CNN         & 861 & 43.0 & 13 & 8.1 \\
SegFormer-B2          & Transformer & \textbf{623} & \textbf{31.1} & 14 & 8.8 \\
\bottomrule
\end{tabular}
\end{table}

\begin{figure}[t]
\centering
\includegraphics[width=\textwidth]{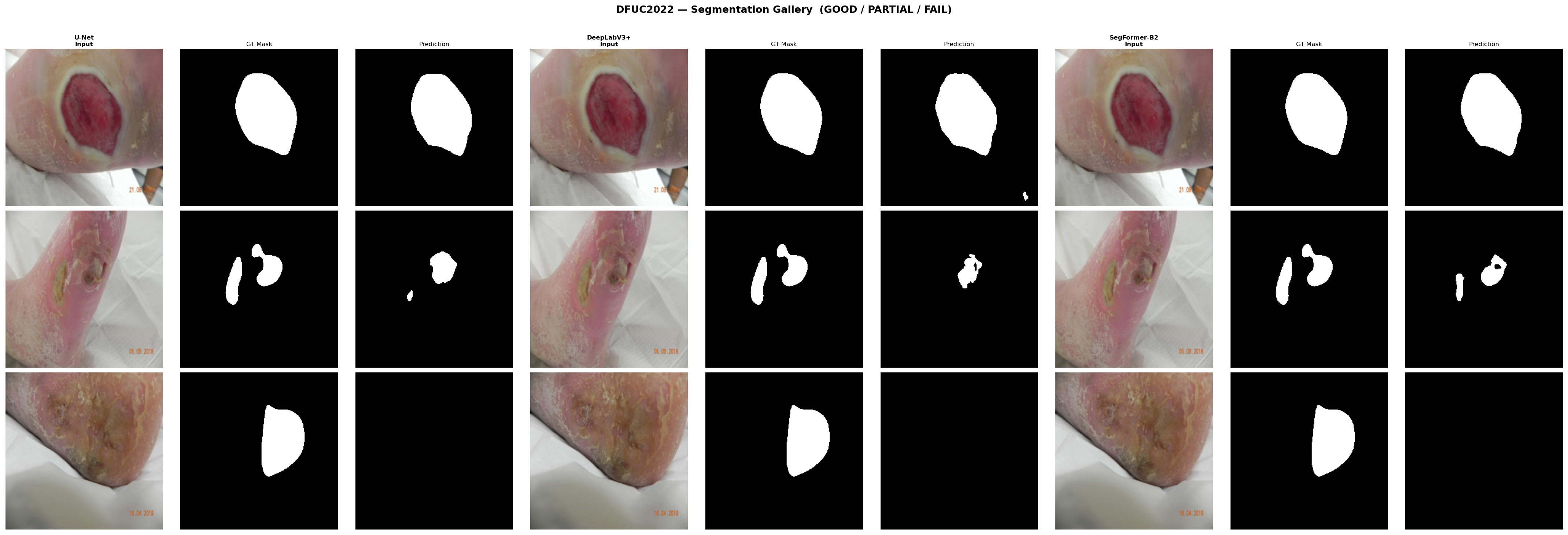}
\caption{Qualitative segmentation results on DFUC2022 for three representative
cases: well-segmented (top), partial (middle), and complete failure (bottom).
For each case: input image, ground-truth mask, and predicted mask are shown per
model. Dice scores are colour-coded: green ($\geq 0.7$), orange (0.4--0.7),
red ($< 0.4$).}
\label{fig:gallery}
\end{figure}

\noindent\textbf{Qualitative analysis.}
Figure~\ref{fig:gallery} illustrates three representative prediction outcomes
on DFUC2022. Good predictions (top row) occur when the wound region is large,
high-contrast, and centrally located. Partial failures (middle row) involve
fragmented or multi-region wounds where models capture only the dominant
region. Complete failures (bottom row) concentrate on small wounds
($<5\%$ of image area), heavily occluded boundaries, or images with low
illumination, in which all models produce near-zero predictions.

\subsection{Statistical Significance}
To confirm that the observed performance differences are not due to chance,
we apply the Wilcoxon signed-rank test (one-sided, paired on per-image Dice).
Table~\ref{tab:wilcoxon} reports all pairwise comparisons.

On DFUC2022, SegFormer-B2 significantly outperforms U-Net
($W = 1{,}018{,}443$, $p < 0.001$) and DeepLabV3+
($W = 1{,}134{,}657$, $p < 0.001$); the CNN ranking is also statistically
significant ($p < 0.001$). On Medetec, SegFormer-B2 significantly
outperforms both U-Net ($W = 9{,}170$, $p < 0.001$) and DeepLabV3+
($W = 9{,}280$, $p < 0.001$), while the difference between the two CNN
models is not significant ($W = 6{,}681$, $p = 0.34$), consistent with
their near-identical mean Dice (0.737 vs.\ 0.730). These results confirm
that the architecture-family advantage of SegFormer-B2 is statistically
reliable across both external test sets.

\begin{table}[t]
\centering
\caption{Wilcoxon signed-rank test results (one-sided, per-image Dice,
$\alpha = 0.05$). *** $p < 0.001$; ns = not significant ($p \geq 0.05$).}
\label{tab:wilcoxon}
\begin{tabular}{llccc}
\toprule
Comparison & Dataset & $W$ & $p$-value & Sig. \\
\midrule
SegFormer-B2 $>$ U-Net      & DFUC2022 & $1{,}018{,}443$ & $8.1\times10^{-21}$ & *** \\
SegFormer-B2 $>$ DeepLabV3+ & DFUC2022 & $1{,}134{,}657$ & $5.5\times10^{-56}$ & *** \\
U-Net $>$ DeepLabV3+        & DFUC2022 & $829{,}160$     & $2.1\times10^{-14}$ & *** \\
SegFormer-B2 $>$ U-Net      & Medetec  & $9{,}170$       & $1.7\times10^{-6}$  & *** \\
SegFormer-B2 $>$ DeepLabV3+ & Medetec  & $9{,}280$       & $6.5\times10^{-7}$  & *** \\
U-Net vs.\ DeepLabV3+       & Medetec  & $6{,}681$       & $0.34$              & ns  \\
\bottomrule
\end{tabular}
\end{table}

\subsection{Discussion}
Three observations emerge from the aggregate and per-image results. First,
the Transformer-based SegFormer not only achieves the best in-domain accuracy
but is consistently the most robust to domain shift across both external test
sets. Its global self-attention mechanism attends to the entire image at every
layer, potentially capturing broader contextual cues---such as surrounding skin
texture, lighting gradients, and wound boundary context---that remain stable
across hospitals, whereas the local receptive fields of convolutional models
may overfit to dataset-specific spatial statistics. Second, complexity alone
does not explain robustness: the heavier DeepLabV3+ generalizes worse than the
simpler U-Net on both external sets, indicating that architecture family
(Transformer vs.\ CNN) is a more decisive factor than parameter count. Third,
the per-image failure analysis, confirmed by Wilcoxon signed-rank tests
($p < 0.001$), reveals that SegFormer-B2's advantage is not driven by
outliers: it produces 238 fewer catastrophic failures than DeepLabV3+ on
DFUC2022, and even on Medetec---where its failure count is marginally
higher---its failures are less severe in magnitude. A key practical implication
is that \emph{high in-domain Dice does not imply deployment readiness}: all
three models achieve strong in-domain scores ($>0.80$) yet lose 28--33 Dice
points when deployed across hospitals. These findings extend prior
cross-dataset DFU analysis~\cite{lucho2024} by showing that modern Transformer
architectures offer measurably better---though still imperfect---generalization.

% =============================================================
\section{Conclusion}
We presented a controlled, leakage-screened cross-dataset benchmark of
convolutional and Transformer architectures for DFU segmentation, augmented
by a per-image failure analysis on $2{,}160$ images across two external test
sets ($2{,}000$ DFUC2022 and 160 Medetec). All models degrade across clinical
sources, but the Transformer-based SegFormer-B2 generalizes best on both
independent external test sets, produces the fewest catastrophic failures on
DFUC2022, and fails less severely than CNN baselines even when it does fail.
All performance rankings are confirmed by Wilcoxon signed-rank tests
($p < 0.001$). Architecture family matters more than model complexity for
cross-hospital robustness. The central practical lesson is that
\textbf{high in-domain Dice does not imply deployment readiness}: a model
scoring $0.83$ on its training distribution may lose a third of that
performance when deployed at a new hospital. A key limitation is that even
the best model loses substantial accuracy across datasets, underscoring that
cross-hospital robustness remains an open problem. Future work includes
domain adaptation, test-time adaptation, and integration into the ILTIAM
longitudinal wound monitoring system for automated wound-area tracking across
clinical visits.

% =============================================================

\end{document}